\documentclass[sigconf]{acmart}
\usepackage{listings}

\lstdefinestyle{Python}{
    language        =   Python, 
    numberstyle     =   \zihao{-5}\ttfamily,
    keywordstyle    =   \color{blue},
    keywordstyle    =   [2] \color{teal},
    stringstyle     =   \color{magenta},
    commentstyle    =   \color{red}\ttfamily,
    breaklines      =   true,   
    columns         =   fixed,  
    basewidth       =   0.5em,
}

\usepackage{cleveref}
\usepackage{hyperref}
\usepackage{multirow}
\usepackage{subcaption}
\usepackage[T1]{fontenc}
\AtBeginDocument{%
  \providecommand\BibTeX{{%
    \normalfont B\kern-0.5em{\scshape i\kern-0.25em b}\kern-0.8em\TeX}}}

\setcopyright{acmcopyright}
\copyrightyear{2020}
\acmYear{2020}
\acmDOI{10.1145/1122445.1122456}

\acmConference[Woodstock '18]{Woodstock '18: ACM Symposium on Neural
  Gaze Detection}{June 03--05, 2018}{Woodstock, NY}
\acmBooktitle{Woodstock '18: ACM Symposium on Neural Gaze Detection,
  June 03--05, 2018, Woodstock, NY}
\acmPrice{15.00}
\acmISBN{978-1-4503-9999-9/18/06}

\begin{document}

\title{XRL-Bench: A Benchmark for Evaluating and Comparing Explainable Reinforcement Learning Techniques}



\author{Yu Xiong$^\dagger$, Zhipeng Hu$^\dagger$, Ye Huang$^\dagger$, Runze Wu$^\dagger$, Kai Guan$^\dagger$, Xingchen Fang$^\dagger$, Ji Jiang$^\dagger$,\\ Tianze Zhou$^\dagger$, Yujing Hu$^\dagger$, Haoyu Liu$^\dagger$,Tangjie Lyu$^\dagger$, Changjie Fan$^\dagger$}

\affiliation{%
    \institution{$^\dagger$Fuxi AI Lab, NetEase Inc. \country{
    Hangzhou, China}}
	\institution{\{xiongyu1,zphu,huangye3,wurunze1,guankai1,fangxingchen,jiangji,zhoutianze,huyujing,\\liuhaoyu03,hzlvtangjie,fanchangjie\}@corp.netease.com}  
}

\renewcommand{\shortauthors}{Yu Xiong et al.}


\begin{abstract}
Reinforcement Learning (RL) has demonstrated substantial potential across diverse fields, yet understanding its decision-making process, especially in real-world scenarios where rationality and safety are paramount, is an ongoing challenge. This paper delves in to Explainable RL (XRL), a subfield of Explainable AI (XAI) aimed at unravelling the complexities of RL models. Our focus rests on state-explaining techniques, a crucial subset within XRL methods, as they reveal the underlying factors influencing an agent's actions at any given time. Despite their significant role, the lack of a unified evaluation framework hinders assessment of their accuracy and effectiveness. To address this, we introduce \textit{XRL-Bench}\footnote{https://github.com/fuxiAIlab/xrl-bench}, a unified standardized benchmark tailored for the evaluation and comparison of XRL methods, encompassing three main modules: standard RL environments, explainers based on state importance, and standard evaluators. XRL-Bench supports both tabular and image data for state explanation. We also propose TabularSHAP, an innovative and competitive XRL method. We demonstrate the practical utility of TabularSHAP in real-world online gaming services and offer an open-source benchmark platform for the straightforward implementation and evaluation of XRL methods. Our contributions facilitate the continued progression of XRL technology.
\end{abstract}



\maketitle

\renewcommand{\thefootnote}{\arabic{footnote}}

\section{Introduction}
Reinforcement Learning (RL) is a popular machine learning paradigm where an agent learns to maximize the accumulated reward through interaction with the environment, and has demonstrated immense potential across various domains, including game AI \cite{ye2020mastering}, robotics \cite{zhao2020sim}, and industrial control systems \cite{nian2020review}. Despite the notable advances in RL, understanding the decision-making process, a critical aspect contributing to its credibility, reliability, and transparency, remains a significant challenge, especially in real-world deployment of RL, where rationality and safety are strongly desired.

Explainable AI (XAI) \cite{vilone2020explainable}, a research field aimed at making the predictive and decision-making process of machine learning models transparent and understandable, has been a growing interest in recently years. Explainable RL (XRL) \cite{vouros2022explainable}, as a subset of XAI, shares the same goal in interpreting RL models. However, due to the complex characteristics of RL, such as its interactivity and dynamism with the environment, the interpretability of XRL is even more challenging. 

A significant category within the realm of XRL methods is the state-explaining techniques. These methods offer intuitive explanations by measuring the importance of state features to an agent's decisions. Extensive research \cite{greydanus2018visualizing, puri2019explain,iyer2018transparency} has been conducted in this particular category, expanding the literature of XAI and paving the way for the application of XAI methodologies in XRL contexts.
However, despite the crucial role these methods play in understanding and explaining decisions made by RL model, challenges similar to those found in the XAI field persist. Specifically, assessing the accuracy and effectiveness of these methods remains a significant challenge due to the absence of a unified standard evaluation framework. This gap hinders the steady and continuous advancement of XRL technologies.


To address this challenge, we propose and develop XRL-Bench, a benchmark for XRL methods. XRL-Bench consists of three main modules: 1) the RL environments, utilizing the open-source game environments; 2) explainers based on state importance, including representative XRL methods and well-known feature attribution method from XAI field; and 3) evaluators that mainly include fidelity and stability measures. Currently, XRL-Bench supports both tabular and image data input for state explanation. This work aims to provide a solid foundation and valuable resource for the ongoing development and evaluation of XRL methods.
The main contributions of this paper are summarized as follows:
\begin{itemize}
    \item We propose a standard and unified framework for evaluating and comparing XRL methods for the first time. This framework is instrumental in understanding the decision-making process of RL models, thereby enhancing their credibility, reliability and transparency.
    \item We introduce a novel XRL method, TabularSHAP, which has demonstrated competitive performance against other XRL methods. Its practical application in real-world services showcased its significant practical value.
    \item We provide an open-source benchmark platform that allows researchers to easily implement and evaluate representative XRL methods through simple APIs. This platform enables the extension of the proposed framework in terms of XRL methods, environements, and evaluation metrics.
\end{itemize}



\section{RELATED WORK}

\subsection{Explainable RL}
Explainable RL attempts to address the interpretable issue by enriching RL models with the ability to provide human-understandable explanations for their actions. Four primary categories exist within XRL approaches, each centered on the explainability of distinct components of the RL framework: model-explaining, reward-explaining, state-explaining and task-explaining methods.

Model-explaining methods focus on extracting the internal logic to create explanations. Depending on the explanation logic type, there are two divisions within these methods: self-explainable and explanation-generating parts. Self-explainable models are designed to be self-explanatory during training, primarily by limiting the model's complexity \cite{du2019techniques, puiutta2020explainable}. Such models can include decision trees \cite{liu2019toward,bastani2018verifiable}, logic rules \cite{jiang2019neural,payani2020incorporating}, or formula expressions \cite{hein2018interpretable}. In contrast, explanation-generating models utilize an auxiliary explicit explanation logic to generate explanations. Examples include counterfactual \cite{olson2021counterfactual,madumal2020explainable} and instruction-based explanations \cite{fukuchi2017autonomous}.


Reward-explaining methods interpret the RL agents' decision-making process by tracking the weight of considered factors in the reward function and determining reasonable reward function weights. These methods can be split into two types: reward decomposition and reward shaping. Reward decomposition \cite{juozapaitis2019explainable,foerster2018counterfactual,wang2020shapley} disassembles the reward function to analyze the influence of its components on the decision-making process and their interrelationships. Reward shaping \cite{mirchandani2021ella,lyu2019sdrl,jin2022creativity}, on the other hand, aims to identify an understandable reward function directly.


State-explaining methods provide explanations based on the environment's states and their influence on the agent's behavior. These methods are critical for understanding an agent's decision-making process, as they shed light on the factors affecting the agent's actions at any given moment. Notable contributions in this category include post-hoc explanation methods like saliency maps \cite{greydanus2018visualizing,puri2019explain}, LIME \cite{ribeiro2016should}, SHAP \cite{lundberg2017unified}, and LRP\cite{bach2015pixel}, as well as attention-based methods \cite{leurent2019social,tang2021sensory}.


Task-explaining methods achieve multi-stage explainability by decomposing the task. The main concept behind Hierarchical Reinforcement Learning (HRL) \cite{barto2003recent} is to create a high-level controller that selects macro-actions and several low-level controllers that choose primitive actions. This division of labor in HRL provides higher architectural explainability by illustrating how a high-level agent schedules low-level tasks. The HRL work can be categorized into whole top-down structure with multi-level task division \cite{shu2017hierarchical,nangue2020boolean} and simple task decomposition with two-level task division \cite{sodhani2021multi,jiang2019language}.

In summary, XRL offers a broad spectrum of explanation methods, each with unique strengths and applications. However, the absence of a unified evaluation framework remains a significant and pressing issue for comparing these XRL methods and generating high-quality explanations \cite{milani2022survey}. To our knowledge, XRL-Bench is the only one that provides standardized environments and evaluation metrics that can be used to assess and compare the quality of explanations generated by various XRL methods. 
The benchmarks and competitions established by our framework hold the potential to propel the advancement of future XRL research.



\subsection{Evaluation Metrics for Explanations}
The significance of evaluation metrics in the design process of XRL systems is widely recognized. Researchers from various disciplines concentrate on distinct objectives of XAI research, which presents difficulties in determining a suitable evaluation methodology. Evaluations of XRL can be broadly categorized into subjective and objective types, depending on the data source for evaluation.

Subjective evaluation assesses explainable frameworks from the human perspective. Key metrics for subjective evaluation can be classified into mental model and user-centric properties. The mental model \cite{vouros2022explainable, yampolskiy2013artificial} pertains to an individual's understanding of the model process, with the explanation facilitating users in building this mental model. One quantitative approach involves allowing testers to predict the agent's decision, and computing the hit rate \cite{ribeiro2016should, ribeiro2018anchors}. User-centric properties such as user satisfaction, trust, and reliance are also considered \cite{lage2019human, gedikli2014should}, reflecting the explanation's complexity, transparency, and utility. However, an excessive reliance on human evaluation may shift the focus towards the persuasiveness of the explanation, neglecting other more abstract aspects such as system transparency, as humans tend to favor simple and effective explanations \cite{herman2017promise}.


Objective evaluation, in contrast, is independent of human assessment. It concentrates on directly measuring the properties of the explainable framework, and can be divided into fidelity, stability, and fairness. Fidelity pertains to the explanation's correlation with the actual rationale for agent decision-making \cite{zhou2021evaluating,liu2021synthetic,agarwal2022openxai}, while stability ensures that the explanations remain largely consistent despite minor perturbations to the input \cite{ghorbani2019interpretation,petsiuk2018rise,alvarez2018robustness}. Fairness ensures the absence of group-based disparities in the fidelity or stability of explanations \cite{dai2022fairness}.  

The effectiveness of subjective evaluation hinges on the competency of the testers, leading to potential variations in the conclusions drawn by testers with different skill levels. Additionally, subjective evaluation might overlook the intrinsic characteristics of the XRL systems and incurs substantial additional time and human resource costs. Hence, our research emphasizes objective evaluation. We have developed XRL-Bench, a framework that focuses on two primary types of objective evaluation: fidelity and stability. Within this framework, we have devised and implemented five persuasive metrics to facilitate the comprehensive evaluation and comparison of various XRL methods.
\section{Overview of 
 XRL-Bench Framework}

\begin{figure*}
    \centering
    \includegraphics[width=0.98\textwidth]{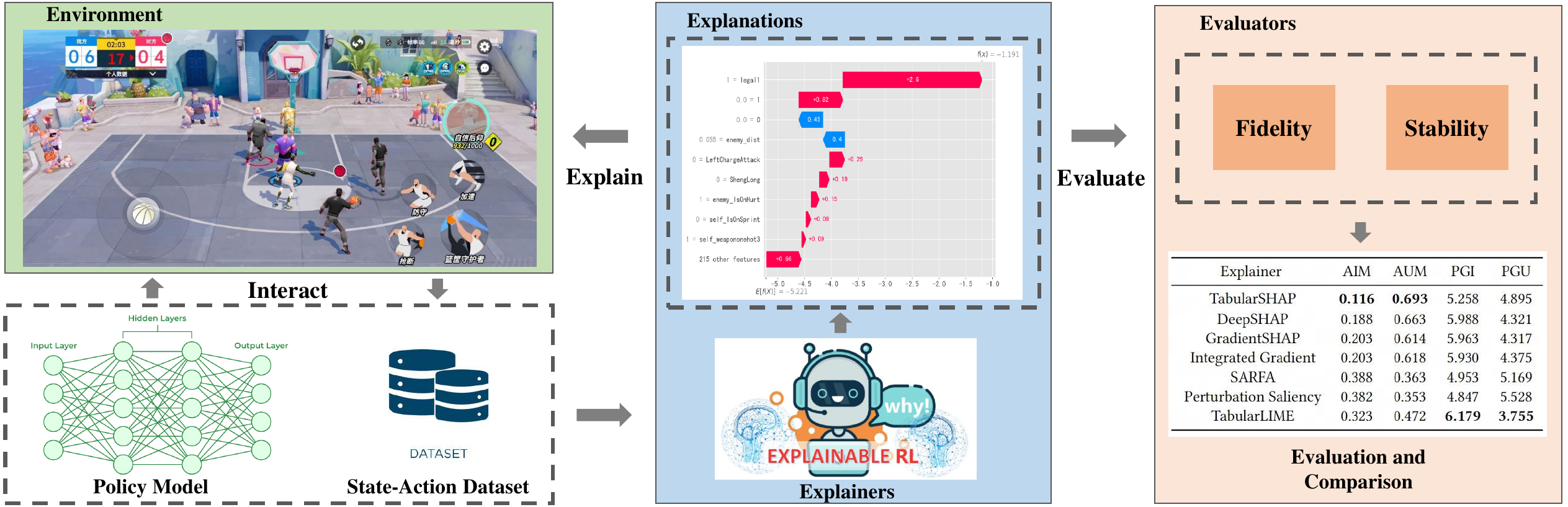}
    \caption{The XRL-Bench framework.}
    \label{fig:framework}
\end{figure*}

The proposed XRL benchmark framework, XRL-Bench, principally comprises three core modules: the RL environments for training policy models and generating interaction datasets, the RL explainers, and the explanation evaluators. This has estabilished a standardized and unified framework for evaluating XRL methods, as illustrated in Figure~\ref{fig:framework}.
XRL-Bench also provides a comprehensive programmatic platform that facilitates researchers and practitioners in the development, testing, and comparative analysis of their state-explaining methods. 


\subsection{Environments, Policy Models and Datasets}
The currently released version of XRL-Bench framework incorporates four tabular input environments and two image input environments. This includes a commercial online basketball game environment, Dunk City Dynasty\footnote{https://www.dunkcitymobile.com/}, developed by NetEase Games, as well as five gym game environments\footnote{https://gymnasium.farama.org/}, all of which are publicly accessible. Each environment features an agent trained using the Deep Reinforcement Learning (DRL) algorithm to accomplish a proficient strategy. The corresponding models and substantial interaction datasets are retained to guarantee reproducibility of all implementations within our XRL-Bench framework.



\textbf{Environments.} Within the tabular state form, we offer four game environments: Dunk City Dynasty, Lunar Lander, Cart Pole, and Flappy Bird. 
The state space of Dunk City Dynasty is a one-dimensional vector of length 520, comprised of the states of six players on the court and the global state.  Lunar Lander's state space dimension is 8, with four potential actions. Cart Pole has a state space dimensionality of four, with two viable actions. Flappy Bird's state space size is 12, with two possible actions.
For the image state form, where each state is depicted by a game screenshot, we provide two game environments: Break Out and Pong. Break Out's state space size is $210*160*3$, with four possible actions, and Pong's state space size matches this, with six potential actions. Our endeavors concentrate on these six standard public game environments, where we train policy models, accumulate interaction data, explain the outcomes, and ultimately evaluate the explanations.



\textbf{Policy Models.} 
For the Dunk City Dynasty environment, we utilize an Actor-Critic algorithm \cite{konda1999actor} with a neural network comprising four fully connected layers with ReLU activation functions for policy learning. Categorical states, such as player ID, are processed through an embedding layer. 
For other tabular input environments, we adopt a DQN algorithm \cite{mnih2013playing} with a neural network consisting of three fully connected layers with ReLU activation functions for policy learning.
Training is concluded once the agent reaches a competitive level, such as when Lunar Lander's most recent 100 episodes average score surpasses 220 points. These policy models are preserved for generating subsequent agent-environment interaction data. For the image states, policy learning also utilizes the DQN algorithm, employing a neural network that includes three two-dimensional convolutional layers and two fully connected layers, all with ReLU activation functions.


\textbf{Datasets.} We generate data via interaction with the environments using policy models and retain it in datasets. For the tabular states, two formats are available: $CSV$ and the $MDPDataset$ from the $d3rlpy$ library. The $CSV$ format affords superior data visualization capabilities, while the $MDPDataset$ format, specifically designed for reinforcement learning, presents a more streamlined structure. For the image states, data is offered in the $MDPDataset$ form. Table~\ref{tab:datasets} provides detailed descriptions about each environment's dataset. It is noteworthy that the state size of the dataset in the image state form is post data preprocessing.

\begin{table}
  \caption{Description of Pre-generated Datasets.}
  \label{tab:datasets}
  \begin{tabular}{ccccc}
    \toprule
    Datasets & State Size & Action Size & S-A Pair Size \\
    \midrule
    \ Dunk City Dynasty & 520 & 52 & 18,889 \\
    \ Lunar Lander & 8 & 4 & 219,392  \\
    \ Cart Pole & 4 & 2 & 125,228 \\
    \ Flappy Bird & 12 & 2 & 129,248 \\
    \ Break Out & $(3, 84, 84)$ & 4 & 3,776  \\
    \ Pong & $(3, 84, 84)$ & 6 & 4,000  \\
    \bottomrule
  \end{tabular}
\end{table}

The XRL-Bench framework provides an abstract $Environment$ class, encapsulating diverse environments while offering a unified and succinct interface. Researchers can conveniently load pre-trained policy models or train their own models based on the selected environment. They can also load pre-generated datasets or create their own. The following code snippet illustrates how to import the Environment class and utilize it to load a dataset corresponding to the environment:

\begin{lstlisting}[
    style = Python,
    caption = {\bf environment.py},
    label = {environment.py}
]
from xrlbench.environments import Environment
environment = Environment(environment_name="lunarLander")
dataset = environment.get_dataset(generate=False)
\end{lstlisting}

\subsection{Explainers}
\label{sec:explianer}
XRL-Bench offers implementations of seven cutting-edge explainable methods for tabular state form. These include TabularSHAP, TabularLIME \cite{ribeiro2016should}, Perturbation Saliency (PS) \cite{greydanus2018visualizing}, SARFA \cite{puri2019explain}, DeepSHAP \cite{lundberg2017unified}, GradientSHAP \cite{erion2019learning}, and Integrated Gradient (IG) \cite{sundararajan2017axiomatic}. Each method will be introduce briefly, with a special focus on the proposed TabularSHAP method. 

TabularLIME, a model-agnostic XAI algorithm, is renowned for generating local explanations by utilizing tabular data. PS formulates saliency maps by applying Gaussian blur perturbations to the state-describing inputs and subsequently gauging the policy changes post information removal. SARFA, another saliency map generating method, differs from PS that are not specific to the action of interest. Instead, SARFA focuses on specificity and relevance.
Specificity measures the impact of perturbations on the Q-value of the action being explained, while relevance downweights features that alter the expected rewards of actions other than the one being explained. DeepSHAP, a SHAP-based adaptation of the original DeepLIFT algorithm \cite{shrikumar2017learning}, is recognized as the fastest neural network explainability approach as it can decompose the output prediction of a neural network on a specific input by backpropagating. IG interpolates between a baseline input and the instance to be explained, calculating gradients at each point and integrating these gradients along the path, which provides a detailed decomposition of the network's predictions. GradientSHAP combines concepts from IG and SHAP to assign importance values to input features, providing a comprehensive understanding of feature contributions.

For the interpretation of image states, XRL-Bench provides implementations of five state-of-the-art explainable methods, which are also applicable to tabular states. These include PS, SARFA, DeepSHAP, GradientSHAP, and IG. Each of these methods have demonstrated significant capabilities in interpreting neural networks.

\textbf{TabularSHAP.} We propose an effective and efficient XRL method TabularSHAP for explaining tabular states. TabularSHAP addresses the challenge of directly interpreting deep neural networks. Initially, it collects interaction data from DRL models and employs ensemble tree models (e.g., LightGBM \cite{ke2017lightgbm}) to learn state-action mapping relationships. As states in tabular form have distinct meanings without strong multi-scale temporal or spatial structures, ensemble tree models are often successful as student models in mimicking the strategies of teacher models. Subsequently, TreeSHAP \cite{lundberg2018consistent}, an XAI method for tree-based models, is employed to interpret the ensemble tree model. TreeSHAP enables the tractable computation of optimal local explanations, known as SHAP values, as defined by desirable properties from the classic game-theoretic Shapley values. By using TreeSHAP to calculate the influence of states on actions, global analysis and episode analysis for RL agents can be formed. This method has been practically applied in DRL-based businesses to provide a clearer understanding of the decision-making process of black-box RL models, and facilitate the quick identification and resolution of issues like unexpected actions. 


XRL-Bench offers an Explainer abstract class to streamline the utilization of various XRL methods. Once researchers have prepared the necessary data or the policy model, they can instantiate the relevant explainer by inputting the desired XRL method name. The process includes a simple preprocessing of the dataset to extract state and action data, the instantiation of the TabularSHAP explainer, and ultimately, the generation of explanations for the state data of interest using the instantiated explainer. The following code snippet illustrates this process:

\begin{lstlisting}[
    style = Python,
    caption = {\bf explainer.py},
    label = {explainer.py}
]
from xrlbench.explainers import Explainer
actions = dataset['action']
states = dataset.drop(['action', 'reward'], axis=1)
explainer = Explainer(method="TabularSHAP", state=states, action=actions)
explanations = explainer.explain(state=states)
\end{lstlisting}




\subsection{Evaluation Metrics}
XRL-Bench incorporates five key quantitative evaluation measures designed to assess the fidelity and stability of XRL methods. For fidelity assessment, the following metrics are provided: Accuracy on Important features Masked by reference padding (AIM), Accuracy on Unimportant features Masked by reference padding (AUM), Prediction Gap on Important feature perturbation (PGI), and Prediction Gap on Unimportant feature perturbation (PGU). Meanwhile, Relative Input Stability (RIS) serves as the metric for stability assessment. These metrics collectively offer a comprehensive evaluation framework for XRL methods.

\textbf{Fidelity.}  The fidelity of an explainer refers to the degree of alignment between the generated explanation and the actual underlying rationale of an agent's decision. High fidelity of an XRL method implies that the produced explanations accurately mirror the true logic driving the agent's decision-making process. We initially use post-hoc accuracy, as introduced in L2X \cite{chen2018learning}, to measure fidelity. This method masks unimportant words, as indicated by the explainer, via zero-padding, and then feeds these into the model for inference. The degree of alignment between the model's inferred categories before and after masking is compared and used to determine the explainer's post-hoc accuracy.

Inspired by this intuitive and applicable XAI evaluation metric, we devised two fidelity measures, namely AIM and AUM. 
AIM masks important states with reference padding values (e.g., zero-padding) and observes the degree of alignment between the model's outputs before and after state masking. 
Conversely, AUM masks unimportant states with reference padding values and observes the degree of disparity between the model's outputs before and after state masking. This is considered as the post-hoc accuracy of the XRL method. 


We also employ PGI and PGU fidelity metrics, as detailed in \cite{petsiuk2018rise,dai2022fairness}. These two metrics compute the difference in prediction value by perturbing important and unimportant states, respectively. For PGI, a higher metric value indicates a higher fidelity of the explanation. A perturbed instance, $x'$, in the local neighborhood of x is generated by slightly perturbing the values of the top-$k$ important states by adding a small amount of Gaussian noise, while keeping all other states constant. Finally, the expected value of the prediction difference between the original and perturbed instances is computed as:

\begin{equation}
    PGI(x,f,e_{x},k)=E_{x'\sim perturb(x,e_{x},top-K)}[|f(x)-f(x')|]
\end{equation}

where $f$ represents the underlying model, and $e_{x}$ signifies an explanation for the model's prediction of $x$. Similar computations are employed for PGU:

\begin{equation}
    PGU(x,f,e_{x},k)=E_{x'\sim perturb(x,e_{x},bottom-K)}[|f(x)-f(x')|]
\end{equation}

\textbf{Stability.} We adopt the evaluation metric presented in \cite{alvarez2018robustness} to assess the robustness of an explanation to minor input perturbations. The Relative Input Stability (RIS) is employed to gauge the maximum alteration in the explanation relative to the input. This evaluation metric can be formalized as follows:


\begin{equation}
RIS(x,x',e_{x},e_{x'})=\underset{x'}{max}\frac{||\frac{e_{x}-e_{x'}}{e_{x}}||_{p}}{max(||\frac{x-x'}{x}||_{p}, \epsilon_{min} )}, \forall x' s.t. x'\in N_{x}
\end{equation}

where $N_{x}$ denotes a neighborhood of instance $x'$ around $x$. The numerator of the metric calculates the $p$-norm of the percentage change of explanation $e_{x'}$ on the perturbed instance $x'$ relative to the explanation $e_{x}$ on the original point $x$. The denominator measures the $p$-norm between the normalized inputs $x$ and $x'$. The maximum term in the denominator safeguards against division by zero.

XRL-Bench also offers an abstract Evaluator class designed for evaluation metrics. By inputting the name of the metric method and the environment, an evaluator instance can be instantiated. This instance allows researchers to swiftly evaluate the generated explanations. The subsequent code snippet illustrates this streamlined process:

\begin{lstlisting}[
    style = Python,
    caption = {\bf evaluator.py},
    label = {evaluator.py}
]
from xrlbench.evaluator import Evaluator
evaluator = Evaluator(metric="AIM", environment=environment)
accuracy = evaluator.evaluate(states, actions, explanations, k=3)
\end{lstlisting}



\textbf{Benchmarking.}
XRL-Bench offers a unified platform that accommodates a broad spectrum of environments, explanation methods, and evaluation metrics. It aids researchers and practitioners in the development, testing, and benchmarking of XRL algorithms. Users can accomplish the entire process, from environment loading, data generation, explainer instantiation, explanation generation, to the final explanation evaluation, with just a few lines of code by constructing instances of the Environment, Explainer, and Evaluator abstract classes. This significantly simplifies and streamlines the development of cutting-edge XRL methods.


\section{Benchmarking Analysis}

Following, we utilize the XRL-Bench framework to perform benchmark tests on the previously discussed state-of-the-art XRL methods. We then proceed with a comparative analysis and discussion of the benchmarking results. All benchmark test experiments are reproducible, with the code repository accessible at the following address: \href{https://github.com/fuxiAIlab/xrl-bench}{https://github.com/fuxiAIlab/xrl-bench}.

\subsection{Experimental Setup}
We executed benchmark tests on tabular states across four distinct environments: $DunkCityDynasty-v1$, $LunarLander-v2$, $CartPole-v0$, and $FlappyBird-v0$. Additionally, for the image states, we assessed XRL methods in two environments, namely $Breakout-v0$ and $Pong-v0$. In XRL-Bench, the reference guidelines for the code implementation of each XRL method are provided within the comments of the respective method class code. 
All XRL methods adhered to the default hyperparameters from their original implementations. For the $DunkCityDynasty-v1$, a four-layer fully connected network with 832, 256, 128, 52 hidden nodes respectively was utilized for the policy training. For other tabular input environments, we use a three-layer fully connected network with 64 hidden nodes. For the image states, the policy network consisted of a three-layer two-dimensional convolutional layer with 32, 64, and 64 convolution kernels respectively, followed by two fully connected layers, with parameters initialized using the Kaiming normal method \cite{he2015delving}. These networks were constructed and trained using the Pytorch framework. For the experimental datasets, trained agents interacted with their corresponding environments over a predetermined number of episodes with a set maximum number of steps, resulting in the pre-generated datasets as depicted in Table~\ref{tab:datasets}.

\subsection{Fidelity}
We conducted an evaluation and comparison of the XRL methods within XRL-Bench, utilizing four fidelity evaluation metrics: AIM, AUM, PGI, and PGU. Initially, it is crucial to comprehend the concept of state importance value as provided by the XRL methods. The question arises: what determines a state's importance? For methods such as SARFA and PS, which produce only positive values, the answer is straightforward - states with higher importance values are deemed more important. However, for other methods that may generate negative values, states with high negative impacts cannot always be disregarded as they often represent critical counterpoints to the model's decisions. Consequently, we define the Top-K states in two ways: 1) The K states with the highest absolute values of state importance; 2) The K states with the highest original values of state importance. Similar definitions are applied to the bottom-K states.
All evaluation methods were calculated according to these two definitions, and the superior results were used for benchmark comparison. The values of the four fidelity evaluation metrics were calculated based on the Area Under the Curve (AUC) over all values of K. For AIM and AUM, zero-padding was employed to mask the most and least important states.

\begin{table*}[]\small
\caption{The fidelity evaluation and comparison of seven XRL methods across four tabular input environments.}
\label{tab:fidelity_tabular}
\begin{tabular}{cllllllllllllllll}
\hline
\multirow{2}{*}{Explainer} & \multicolumn{4}{c}{Dunk City Dynasty}                                                  & \multicolumn{4}{c}{Lunar Lander}                                                       & \multicolumn{4}{c}{Cart Pole}                                                          & \multicolumn{4}{c}{Flappy Bird}                                   \\ \cline{2-17} 
                           & AIM$\downarrow$            & AUM$\uparrow$            & PGI$\uparrow$            & \multicolumn{1}{l|}{PGU$\downarrow$}            & AIM$\downarrow$            & AUM$\uparrow$            & PGI$\uparrow$            & \multicolumn{1}{l|}{PGU$\downarrow$}            & AIM$\downarrow$            & AUM$\uparrow$            & PGI$\uparrow$            & \multicolumn{1}{l|}{PGU$\downarrow$}            & AIM$\downarrow$            & AUM$\uparrow$            & PGI$\uparrow$            & PGU$\downarrow$            \\ \hline
TabularSAHP                & \textbf{0.214} & \textbf{0.894} & 0.905          & \multicolumn{1}{l|}{0.662}          & \textbf{0.116} & \textbf{0.693} & 5.258          & \multicolumn{1}{l|}{4.895}          & \textbf{0.134} & \textbf{0.960} & 0.452          & \multicolumn{1}{l|}{0.522}          & \textbf{0.331} & 0.566          & 3.019          & \textbf{1.685} \\
DeepSHAP                   & 0.337          & 0.493          & 0.790          & \multicolumn{1}{l|}{0.712}          & 0.188          & 0.663          & 5.988          & \multicolumn{1}{l|}{4.321}          & 0.377          & 0.740          & 0.651          & \multicolumn{1}{l|}{0.324}          & 0.542          & 0.472          & 2.970          & 1.759          \\
GradientSHAP               & 0.326          & 0.523          & 0.766          & \multicolumn{1}{l|}{0.690}          & 0.203          & 0.614          & 5.963          & \multicolumn{1}{l|}{4.317}          & 0.268          & 0.827          & \textbf{0.655} & \multicolumn{1}{l|}{0.312}          & 0.631          & 0.517          & 2.953          & 1.792          \\
IG                         & 0.323          & 0.522          & 0.808          & \multicolumn{1}{l|}{0.688}          & 0.203          & 0.618          & 5.930          & \multicolumn{1}{l|}{4.375}          & 0.260          & 0.837          & 0.654          & \multicolumn{1}{l|}{\textbf{0.311}} & 0.609          & 0.511          & \textbf{3.104} & 1.738          \\
SARFA                      & 0.361          & 0.709          & 0.952          & \multicolumn{1}{l|}{0.665}          & 0.388          & 0.363          & 4.953          & \multicolumn{1}{l|}{5.169}          & 0.602          & 0.558          & 0.503          & \multicolumn{1}{l|}{0.569}          & 0.593          & \textbf{0.573} & 2.598          & 2.671          \\
PS                         & 0.364          & 0.687          & 0.951          & \multicolumn{1}{l|}{0.680}          & 0.382          & 0.353          & 4.847          & \multicolumn{1}{l|}{5.528}          & 0.556          & 0.574          & 0.411          & \multicolumn{1}{l|}{0.569}          & 0.584          & 0.565          & 2.568          & 2.639          \\
TabularLIME                & 0.215          & 0.779          & \textbf{0.954} & \multicolumn{1}{l|}{\textbf{0.274}} & 0.323          & 0.472          & \textbf{6.179} & \multicolumn{1}{l|}{\textbf{3.755}} & 0.613          & 0.564          & 0.646          & \multicolumn{1}{l|}{0.329}          & 0.559          & 0.511          & 3.017          & 2.038          \\ \hline
\end{tabular}
\end{table*}

\textbf{Comparison in Tabular State Form.} 
The comparative analysis of XRL methods in tabular state form is displayed in Tables~\ref{tab:fidelity_tabular}. The results demonstrate that TabularSHAP surpasses other methods in AIM and AUM metrics, registering the top results in 7 out of 8 evaluations across three data sets. This underscores TabularSHAP's superior capacity to select important states, suggesting that the bulk of RL model strategies can be replicated by focusing on a minority of critical states. Notably, in the Flappy Bird dataset, characterized by an imbalance where the ratio of $do nothing$ to $flap$ actions approximates $10:1$, all other methods were unable to accurately reconstruct model decisions using their important states. Their AUC on AUM consistently surpassed that on AIM. Only TabularSHAP was able to successfully identify the important states under these challenging conditions. TabularSHAP's performance on PGI and PGU metrics is average, which we attribute to the uncertainty introduced by perturbations. The perturbation concept lacks a solid theoretical framework and often produces results that lack precision in practical application. The intricate decision boundaries of complex models in high-dimensional data, with perturbations in varying directions and scales, lead to diverse effects. The subpar performance observed in perturbation-based methods such as SARFA, PS, and TabularLIME further underscores the limitations of perturbation-based approaches. 

DeepSHAP, GradientSHAP, and IG display similar and competitive performances, all leveraging gradients to calculate state importance. These methods achieved the top results in 3 out of 8 PGI and PGU metrics across three datasets, indicating that gradient-based methods excel at capturing the variance in the original RL model's outputs after state perturbations. Despite SARFA and PS also computing state importance based on the perturbation concept, they differ from the perturbations in PGI and PGU metrics. PGI and PGU focus solely on the prediction result differences in the target action, while SARFA and PS concentrate on the prediction result differences across all actions, leading to their less stable fidelity performance. In general, TabularLIME exhibits moderate performance, but it demonstrates exceptional results in PGI and PGU, achieving top performance in four evaluations. This implies that promising results are likely when the target action of interest for perturbation in the method coincides with that of the metrics.




\begin{table}[]
\caption{The fidelity evaluation and comparison of five XRL methods across two image input environments.}
\label{tab:fidelity_image}
\begin{tabular}{cllll}
\hline
\multirow{2}{*}{Explainer} & \multicolumn{4}{c}{Break Out}                                     \\ \cline{2-5} 
                           & AIM$\downarrow$ & AUM$\uparrow$ & PGI$\uparrow$ & PGU$\downarrow$            \\ \hline
DeepSHAP                   & \textbf{0.162} & 0.630          & 1.748          & \textbf{0.347} \\
GradientSHAP               & 0.260          & \textbf{0.655} & 1.755          & 0.384          \\
IG                         & 0.292          & 0.652          & \textbf{1.812} & 0.364          \\
SARFA                      & 0.253          & 0.270          & 1.225          & 0.991          \\
PS                         & 0.258          & 0.387          & 1.370          & 0.621          \\ \hline
\multirow{2}{*}{Explainer} & \multicolumn{4}{c}{Pong}                                          \\ \cline{2-5} 
                           & AIM$\downarrow$ & AUM$\uparrow$ & PGI$\uparrow$ & PGU$\downarrow$            \\ \hline
DeepSHAP                   & \textbf{0.111} & 0.160          & 2.271          & \textbf{0.699} \\
GradientSHAP               & 0.114          & 0.153          & \textbf{2.453} & 1.164          \\
IG                         & 0.128          & \textbf{0.169} & 2.213          & 0.770          \\
SARFA                      & 0.147          & 0.148          & 2.375          & 1.500          \\
PS                         & 0.148          & 0.15           & 2.346          & 1.077          \\ \hline
\end{tabular}
\end{table}



\textbf{Comparison in Image State Form.} Given the extensive state space associated with image input, calculating the AUC over all K values is computationally demanding. To facilitate a more efficient evaluation and comparison of XRL methods, we compute the AUC for K mod 10. The comparative results are displayed in Table~\ref{tab:fidelity_image}. Based on the evaluation results of AIM and AUM across two environments, DeepSHAP and GradientSHAP, especially the former, exhibit superior performance. This underscores the robust theoretical foundation of SHAP-based methods, with IG coming next, thus emphasizing the effectiveness of gradient-based methods in interpreting neural networks. A similar conclusion can be drawn from the PGI and PGU evaluation results, namely that the fidelity of gradient-based explainable methods significantly outperforms that of perturbation-based explainable methods, further indicating that the theoretical foundation of the latter requires bolstering.

\subsection{Stability}

\begin{table*}[]
\caption{The RIS stability evaluation and comparison of the XRL methods across four tabular input environments and two image input environments.}
\label{tab:ris}
\begin{tabular}{ccccccc}
\hline
Explainer    & Dunk City Dynasty & Lunar Lander   & Cart Pole       & Flappy Bird    & Break Out        & Pong             \\ \hline
TabularSHAP  & \textbf{1.023}             & 2.646          & 30.734          & 11.403         & \textbackslash{} & \textbackslash{} \\
DeepSHAP     & 1.261             & \textbf{2.623} & 21.104          & \textbf{7.355} & 0.375            & \textbf{0.044}   \\
GradientSHAP & 1.419             & 3.134          & 18.520          & 10.732         & 0.659            & 5.490            \\
IG           & 1.465             & 2.800          & \textbf{13.713} & 10.339         & \textbf{0.109}   & 0.758            \\
SARFA        & 1.071    & 6.408          & 54.193          & 10.419         & 0.653            & 2.291            \\
PS           & 1.357             & 4.878          & 54.694          & 11.940         & 0.649            & 3.570            \\
TabularLIME  & 1.901             & 2.871          & 34.063          & 14.691         & \textbackslash{} & \textbackslash{} \\ \hline
\end{tabular}
\end{table*}

We evaluated and compared the stability of seven XRL methods for tabular states across four environments, as shown in Table~\ref{tab:ris}.
The resutls reveal that DeepSHAP 
yielded the most stable explanations, followed by IG and GradientSHAP. This demonstrates the efficacy of gradient-based methods in explaining neural networks, with gradient integration serving to enhance the smoothness of the process. Despite being an indirect interpretation of the RL model, TabularSHAP shows commendable stability. Coupled with its superior performance in fidelity, this reflects the feasibility of the approach of indirectly explaining RL models. On the other hand, the stability performance of perturbation-based explainable methods was unsatisfactory.
We further evaluated the stability of five XRL methods for image states across two environments, as presented in Table~\ref{tab:ris}. DeepSHAP and IG again achieves the best stability results, indicating that these two methods can maintain considerable stability across different data formats. The stability performance of the remaining methods was relatively mediocre in general.





\subsection{Computational Efficiency}

\begin{figure}
    \centering
    \includegraphics[width=0.42\textwidth]{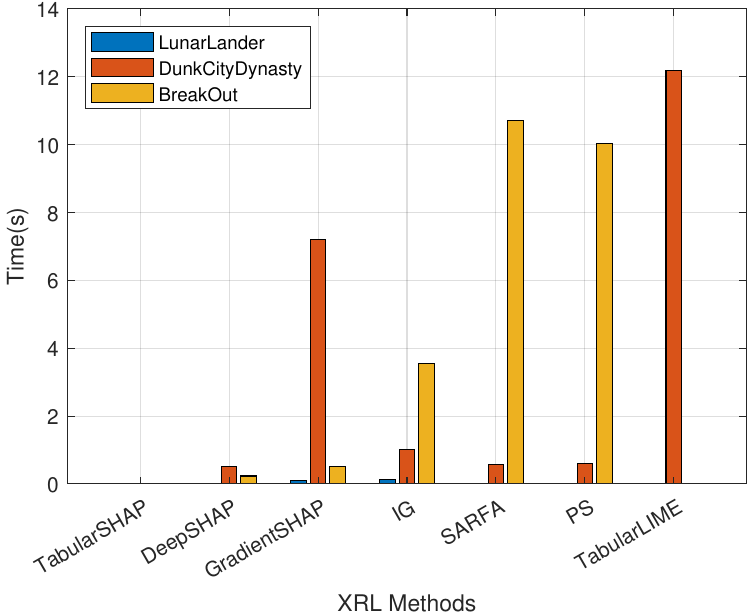}
    \caption{The computational efficiency comparison of seven XRL methods.}
    \label{fig:efficiency}
\end{figure}

The generation of explanations inherently requires additional time.
An XRL methods with higher efficiency can produce explanations more promptly, which often enhances its practical utility. We conducted comparative experiments on the computational efficiency of seven XRL methods on a PC with an Intel Core i9 3.6 GHz 16-core processor. Figure~\ref{fig:efficiency} presents the time taken to generate explanations for each sample. In the Lunar Lander environment, which has a small state space, all seven methods took less than 0.15s to compute an explanation. In the more complex game environment, Dunk City Dynasty, which features a larger state space, TabularSHAP demonstrated a clear advantage of efficiency, requiring only 0.003s per sample. This is particularly beneficial for applications that handle large-scale data and need to provide explanations promptly. Conversely, GradientSHAP and TabularLIME were less efficient, with explanation times averaging around 10 seconds per sample. In the image input environment of Break Out, DeepSHAP outperformed the other four methods, generating explanations for each sample in just 0.234 seconds.


\section{Real-World Application of XRL: A Case Study}
Despite the proliferation of XRL methods in recent years, their effective deployment in practical environments remains a challenge due to the inefficiencies, ineffectiveness, or obscurity of the generated explanations. This section presents a case study where we successfully apply XRL methods in a RL AI project focused on online gaming. We aim to assist practitioners in narrowing the divide between theoretical algorithms and their practical implementations, thereby fostering further research interest in the XRL field.


\subsection{Role of AI Bots in Online Gaming}
Online gaming necessitate a variety of AI components, including non-player characters (NPCs), boss monsters, and matchmaking bots. These elements contribute to the realism and diversity of the gaming world. However, relying on traditional developers to manually encode behavior logic often leads to subpar performance for simple AI, high development costs for complex AI, and issues such as predictable actions, repetitive patterns, and difficulties in managing complex scenarios. By applying RL techniques, which enable algorithms to continually learn AI control, these problems can be effectively mitigated. 

RL AI bots, developed by NetEase, have been successfully integrated into various games, proving advantageous in player versus environment (PvE) and player versus player (PvP) scenarios and meeting a variety of player needs. However, the intricate and opaque structure of RL models can result in inefficient debugging processes during model training. In practical applications, when an AI's behavior does not meet human expectations, engineers typically follow a three-step process of problem assumption, experimental verification, and adjustment improvement. This process often requires numerous iterations, is time-consuming, and inefficient due to the slow training nature of RL models. It also heavily depends on the engineer's experience. Consequently, understanding the decision-making mechanism of the RL model and rendering AI behavior intelligible to humans is crucial for expediting problem identification and accelerating AI bot development.

\subsection{Methodology for XRL Implementation}
To expedite the analysis of AI bot behavior patterns in online gaming, and to promptly identify and rectify anomalous actions, we devised and executed a precise, efficient, and pragmatic XRL strategy. This approach, rooted in the TabularSHAP method detailed in Section~\ref{sec:explianer}, is utilized to interpret the decision-making mechanism of RL models. The strategy unfolds in four distinct phases:
\begin{enumerate}
    \item \textbf{Collection of State-Action Data }: We select the RL model for interpretation, execute combat games repeatedly, and collect the ensuing state-action data pairs. This creates a state-action dataset of size N, which is employed for subsequent explanation analysis.    
    \item \textbf{Estimation of State-to-Action Influence}: We utilize the TabularSHAP method to calculate the state factors that instigate each action, along with their respective SHAP values.
    \item \textbf{Global analysis}: The computed SHAP values are graphically represented using Summary and Dependence Plots, as depicted in Figure~\ref{fig:summaryplot} and Figure~\ref{fig:dependenceplot}. This allows engineers to quickly understand the broader picture of the factors influencing the AI bot's behavior, and further scrutinize any outcomes that do not conform to human expectations.
    \item \textbf{Episode analysis}: For an episode of anomalous bot behavior, we compute the SHAP values of the state-to-action within the episode data using TabularSHAP. For an anomalous action at a specific timestep, we employ the Waterfall Plot, as seen in Figure~\ref{fig:waterfallplot}, to illustrate the key state factors that influence the occurrence of the action. Simultaneously, we delve deeper into the significant factors of an action that was anticipated but did not occur, identifying the core factors that inhibited the activation of the action. Through these factual and counterfactual analysis, we can swiftly identify the anomalous state, providing a strong foundation for troubleshooting and improving the efficiency of problem-solving.
\end{enumerate}

\subsection{Case Study}
\begin{figure}
    \centering
    \includegraphics[width=0.48\textwidth]{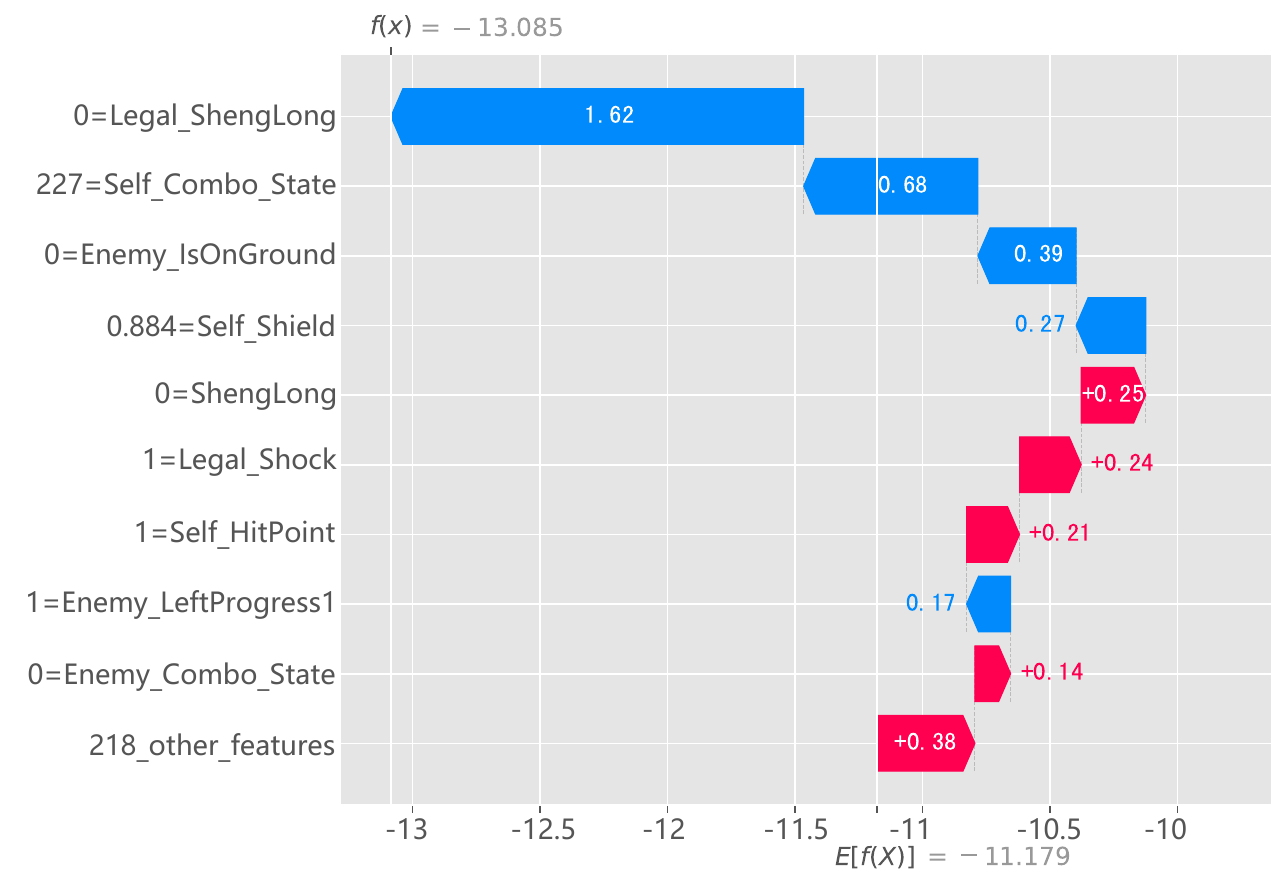}
    \caption{The Waterfall plot for XRL episode analysis. Waterfall plot demonstrates each state's contribution in pushing the model output from its base value (the average model output over the dataset) to the final model output. States that increase the model prediction are depicted in red, while those that decrease it are in blue.}
    \label{fig:waterfallplot}
\end{figure}

\begin{figure}[h]
    \centering
    \includegraphics[width=0.48\textwidth]{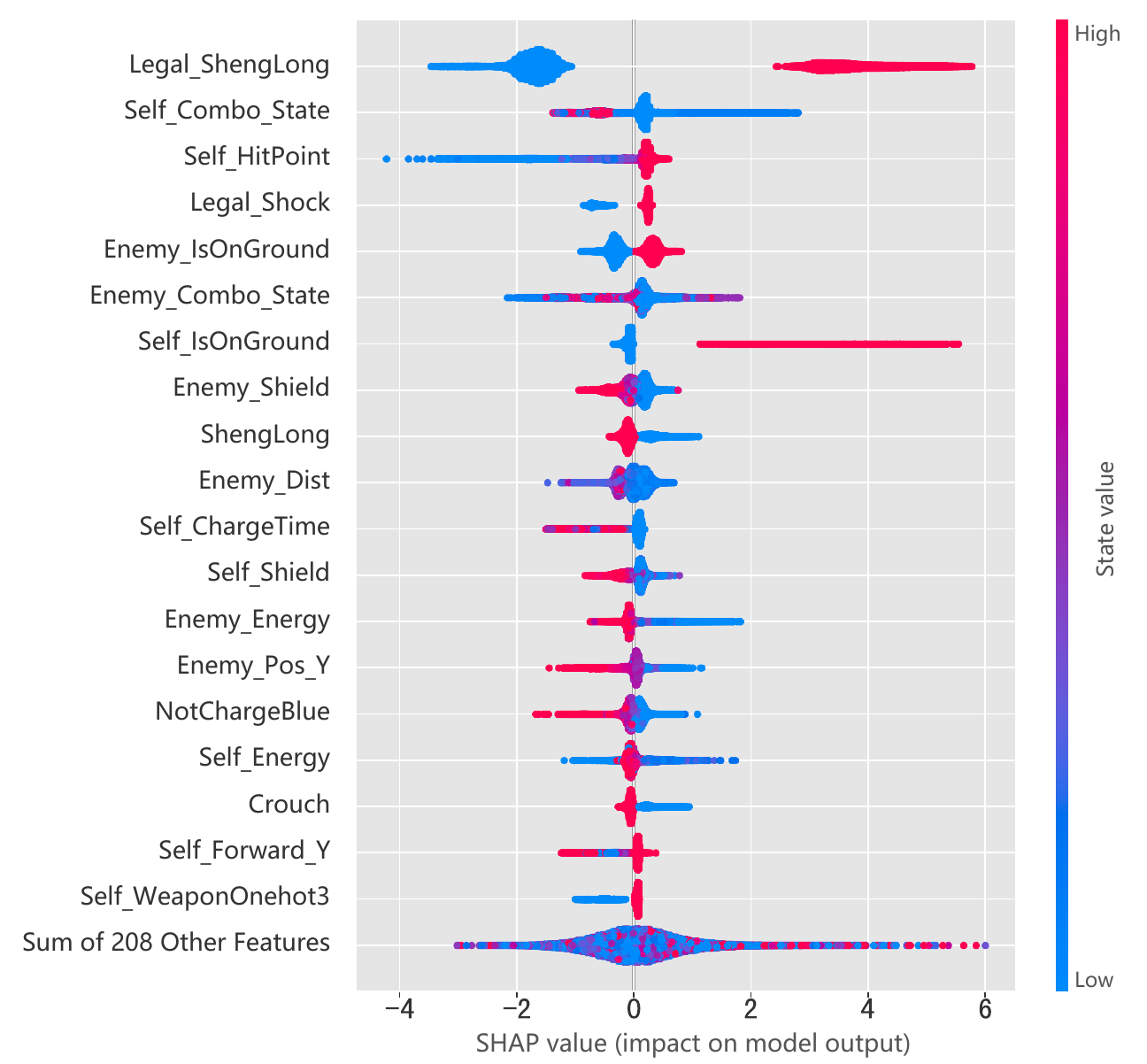}
    \caption{The Summary plot for XRL global analysis. Summary plot organizes states based on the cumulative magnitude of their SHAP values and uses these values to depict the distribution of each state's influence.}
    \label{fig:summaryplot}
\end{figure}

\begin{figure}[h]
    \centering
    \includegraphics[width=0.45\textwidth]{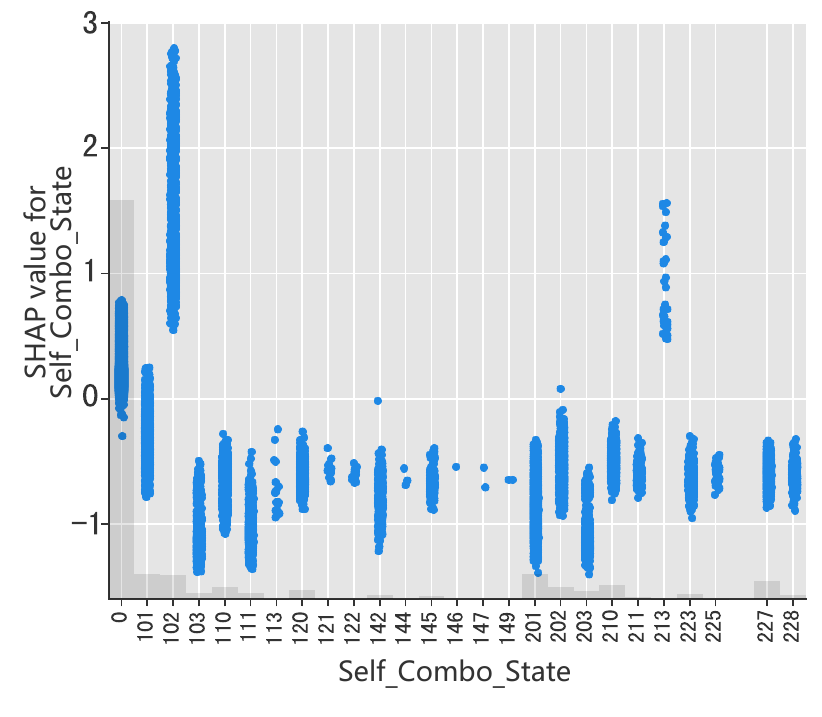}
    \caption{The Dependence plot for XRL global analysis. Dependence plot uses the SHAP value of a specific state as the y-axis, while the corresponding feature value is represented on the x-axis.}
    \label{fig:dependenceplot}
\end{figure}

The application case\footnote{https://github.com/fuxiAIlab/xrl-bench/tree/main/video} of XRL comes from \textit{Naraka:Bladepoint}\footnote{https://www.narakathegame.com/}, a highly sought after multiplayer action-competitive game by NetEase Games. In situations where AI bots struggle to seamlessly connect the $LeftAttack$ to the $ShengLong$ skill - a move considered expert routine. Instead, the AI bots tend to perform $LeftAttack$ twice before transitioning to the $ShengLong$ skill. Initially, a global analysis is conducted to scrutinize the principal factors that trigger the activation of the $ShengLong$ skill. As depicted in Figure~\ref{fig:summaryplot} and Figure~\ref{fig:dependenceplot}, the top five factors are the availability of the $ShengLong$ skill, the player's state in the combo chart, the player's high hit point, the availability of the $Shock$ skill, and the opponent's grounded state. These analytical results coincide with human intuition, validating the overall normal functioning of the RL model. Subsequently, an episode analysis is conducted on a dataset that exhibits the aforementioned anomaly. We zero in on the segment where $LeftAttack$ is executed at the 10th timestep, followed by another $LeftAttack$ at the 11th timestep, and finally $ShengLong$ at the 12th timestep. The focus is on understanding the rationale behind executing $LeftAttack$ instead of $ShengLong$ at the 11th timestep. The analysis, as illustrated in Figure~\ref{fig:waterfallplot}, reveals that the primary reason for not executing $ShengLong$ is its unavailability (legal is 0). Upon further investigation, it was found that a programming interface bug was responsible for this issue. The AI bot, after executing an attack, was not recognized as still being on the ground, leading to a false legal status for $ShengLong$. Once this bug was rectified, the problem was resolved, and the AI bots successfully learned to connect $LeftAttack$ to $ShengLong$, mastering the expert routine.

\section{Conclusion}

This paper has advanced the field  of XRL by proposing XRL-Bench, a benchmarking framework for XRL methodologies. The primary challenge that this work addresses is the lack of a unified evaluation framework for XRL techniques, which has been a significant obstacle to the consistent progression of XRL technologies. Our solution, XRL-Bench, encompasses three main modules: RL environments, explainers based on state importance, and evaluators which primarily include fidelity and stability measures. It supports both tabular and image data input for state explanation, thereby extending its applicability across various RL scenarios. In addition, we introduced a novel XRL method, TabularSHAP, which has demonstrated competitive performance against other XRL methods. Its practical application in real-world online gaming services showcased its practical relevance and received wide recognition. Furthermore, we have provided an open-source benchmark platform that allows researchers and practitioners to easily implement and evaluate representative XRL methods through simple APIs. This platform enables the extension of the proposed framework in terms of XRL methods, environments, and evaluation metrics. Overall, this work contributes a solid foundation and a valuable resource for the ongoing development and evaluation of XRL methods, thereby paving the way for further advancements in this crucial field of study.


\balance
\bibliographystyle{ACM-Reference-Format}
\bibliography{main}

\section{Supplement}

Figures 4 and 5 present the Summary Plot and Dependence Plot, respectively, of the application of XRL technology in the case of AI bots for online gaming. These analyses serve to provide a comprehensive understanding of the global influence from state to action. Developers of AI bots can utilize this global analysis to rapidly comprehend the overview of the decision-making mechanism of the model.


\end{document}